\documentclass[12pt,twoside]{article}   
\usepackage[super,sort,comma]{natbib}

\usepackage{fancyhdr}		

\usepackage[section]{placeins}   %

\usepackage{graphicx}
\usepackage{amsmath,amssymb,amsfonts}
\usepackage{booktabs}
\usepackage{multirow}
\usepackage{makecell}
\usepackage{textcomp}

\makeatletter \renewcommand\@biblabel[1]{$^{#1}$} \makeatother
\newcommand{\cen}[1]{\begin{center} #1 \end{center}}

\usepackage[mathlines]{lineno}

\usepackage{soul}
\soulregister\cite7
\soulregister\citep7
\soulregister\citet7
\soulregister\ref7
\soulregister\pageref7

\usepackage{hyperref}
\hypersetup{ colorlinks,
    citecolor=blue,
    filecolor=blue,
    linkcolor=blue,
    urlcolor=blue
}

\usepackage{xcolor}

\definecolor{gray}{rgb}{0.6,0.6,0.6}
\definecolor{red}{rgb}{0.85,0,0}
\definecolor{green}{rgb}{0,0.85,0}
\definecolor{blue}{rgb}{0,0,0.85}
\definecolor{beige}{rgb}{0.92,0.87,0.78}
\usepackage[all]{hypcap}    

\begin{document}

\cen{\sf {\Large {\bfseries Decoupled Pyramid Correlation Network for Liver Tumor Segmentation from CT images} \\  
\vspace*{10mm}
Yao Zhang\textsuperscript{1,2}, Jiawei Yang\textsuperscript{3}, Yang Liu\textsuperscript{1,2}, Jiang Tian\textsuperscript{5}, Siyun Wang\textsuperscript{4}, Cheng Zhong\textsuperscript{5}, Zhongchao Shi\textsuperscript{5}, Yang Zhang\textsuperscript{6}, Zhiqiang He\textsuperscript{7}} \\
\textsuperscript{1}Institute of Computing Technology, Chinese Academy of Sciences, Beijing, China
\\
\textsuperscript{2}University of Chinese Academy of Sciences, Beijing, China\\
\textsuperscript{3}Electrical and Computer Engineering, University of California, Los Angeles, U.S.\\
\textsuperscript{4}Dornsife College of Letters, Arts and Sciences, University of Southern California, Los Angeles, U.S.\\
\textsuperscript{5}AI Lab, Lenovo Research, Beijing, China\\
\textsuperscript{6}Lenovo Research, Beijing, China\\
\textsuperscript{7}Lenovo Ltd., Beijing, China
\vspace{5mm}\\
Version typeset \today\\
}

\pagenumbering{roman}
\setcounter{page}{1}
\pagestyle{plain}
Zhiqiang He and Yang Zhang are the corressponding authors. email: hezq@lenovo.com and zhangyang20@lenovo.com. This work is done when Yao Zhang works as an intern at AI Lab, Lenovo Research. \\

\begin{abstract}
\noindent {\bf Purpose:} Automated liver tumor segmentation from Computed Tomography (CT) images is a necessary prerequisite in the interventions of hepatic abnormalities and surgery planning. However, accurate liver tumor segmentation remains challenging due to the large variability of tumor sizes and inhomogeneous texture. Recent advances based on Fully Convolutional Network (FCN) for medical image segmentation drew on the success of learning discriminative pyramid features. In this paper, we propose a Decoupled Pyramid Correlation Network (DPC-Net) that exploits attention mechanisms to fully leverage both low- and high-level features embedded in FCN to segment liver tumor.\\
{\bf Methods:} We first design a powerful Pyramid Feature Encoder (PFE) to extract multi-level features from input images. Then we decouple the characteristics of features concerning spatial dimension (i.e., height, width, depth) and semantic dimension (i.e., channel). On top of that, we present two types of attention modules, Spatial Correlation (SpaCor) and Semantic Correlation (SemCor) modules, to recursively measure the correlation of multi-level features. The former selectively emphasizes global semantic information in low-level features with the guidance of high-level ones. The latter adaptively enhance spatial details in high-level features with the guidance of low-level ones.\\
{\bf Results:} We evaluate the DPC-Net on MICCAI 2017 LiTS Liver Tumor Segmentation (LiTS) challenge dataset. Dice Similarity Coefficient (DSC) and Average Symmetric Surface Distance (ASSD) are employed for evaluation. The proposed method obtains a DSC of 76.4\% and an ASSD of 0.838 mm for liver tumor segmentation, outperforming the state-of-the-art methods. It also achieves a competitive results with a DSC of 96.0\% and an ASSD of 1.636 mm for liver segmentation. \\
{\bf Conclusions:} The experimental results show promising performance of DPC-Net for liver and tumor segmentation from CT images. Furthermore, the proposed SemCor and SpaCor can effectively model the multi-level correlation from both semantic and spatial dimensions. The proposed attention modules are lightweight and can be easily extended to other multi-level methods in an end-to-end manner. 

\end{abstract}

\newpage     

\tableofcontents

\newpage

\setlength{\baselineskip}{0.7cm}      

\pagenumbering{arabic}
\setcounter{page}{1}
\pagestyle{fancy}
\section{Introduction}

\begin{figure}[!tp]
\centering
\includegraphics[width=0.5\linewidth]{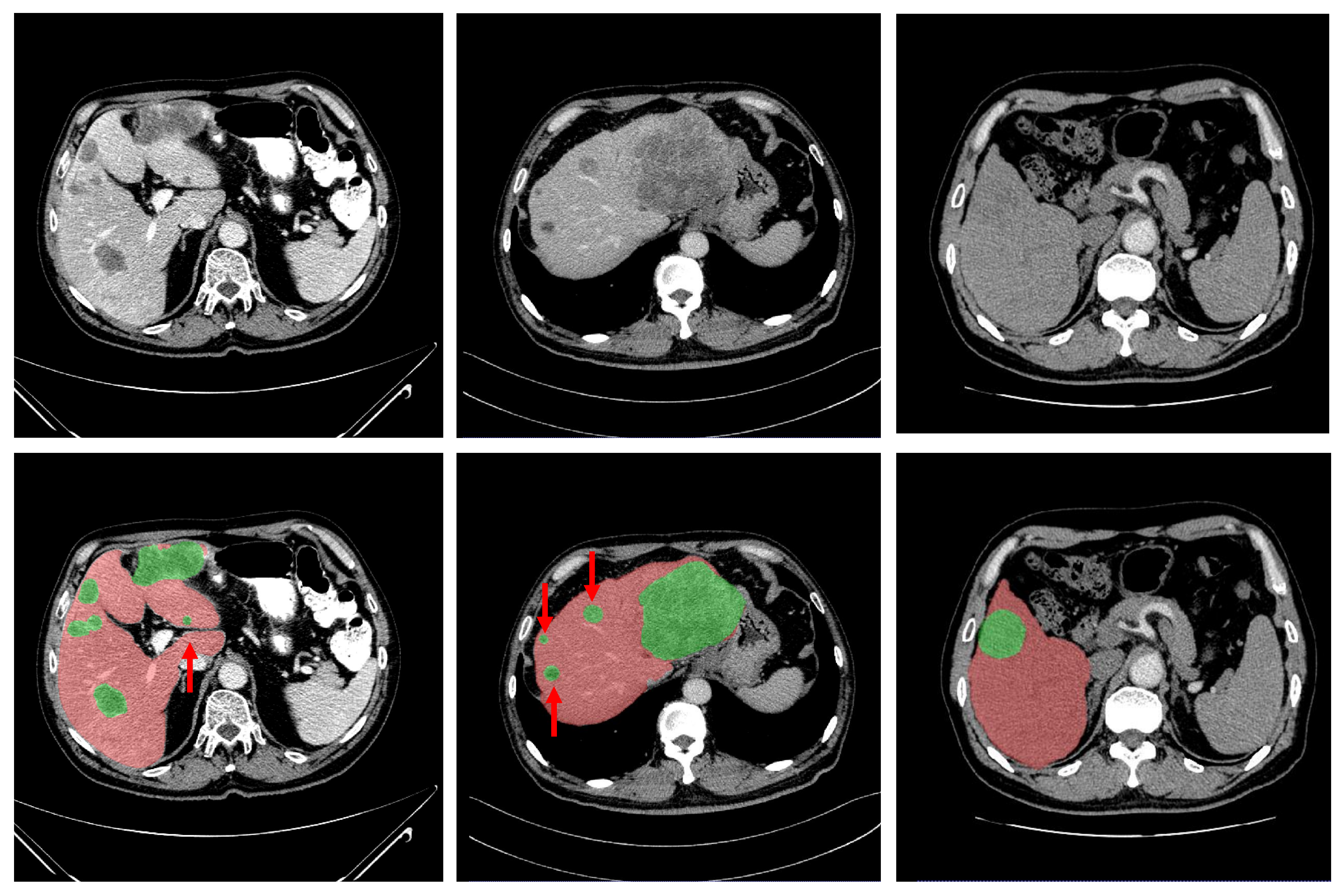}
\caption{Examples of contrast-enhanced CT scans showing various sizes, shapes, locations, and textures of liver tumors within patients. 
The first row is the CT scans, and the second row is the ground truth. The red regions denote the liver while the green ones denote the tumors. Red arrows highlight small tumors.
}
\label{fig:examples}
\end{figure}

Liver and tumor segmentation from Computed Tomography (CT) is a mandatory task in diagnosing, monitoring, and treating liver diseases. It provides accurate measurements of the shape, volume, and location of organs, tissues, tumors, lesions, and other anatomical structures. This biomedical information is crucial to assist doctors in making evaluations and planning for surgeries. Manual annotation of medical images is tedious and time-consuming, requiring professional knowledge and skill. Moreover, it is prone to inter-operator and intra-operator variation, which makes the procedure non-reproducible. Therefore, automated liver and tumor segmentation is highly demanded in clinical practice for efficiency and reproducibility. Compared with liver segmentation, liver tumor segmentation is considered a much more challenging task. Firstly, a certain number of liver tumors share similar textures or have unclear boundaries, making them indistinguishable in semantic space. Secondly, liver tumors within the patients usually have various sizes, shapes, and locations, making the model confused in spatial space. These problems pose difficulties for both data annotation and segmentation. Figure~\ref{fig:examples} illustrates some typical examples of liver tumors in CT volumes.

It is fundamental to build pyramid features, i.e., multi-level features, to recognize objects with various sizes, shapes, and locations~\cite{adelson1984pyramid}. Please note that pyramid features here do not only denote features with different resolutions but also receptive fields and contextual information.
In recent years, semantic segmentation has witnessed unprecedented progress obliged to the great success of Fully Convolutonal Network (FCN)~\cite{long2015fully}, which can generate segmentation in an end-to-end manner by extracting multi-level features through convolutional and pooling layers. Methods in this stream could be broadly classified into two categories: backbone-based and encoder-decoder methods. Backbone-based methods are built upon a deep backbone network with or without simple up-sampling operations. 
The segmentation is generated at the top of the backbone without explicitly exploiting the low-level features in shallow layers, which restricts the feature reuse and thus results in inefficient representation~\cite{huang2017densely}.
Encoder-decoder methods employ a decoder, a counterpart of the encoder, and construct symmetric architecture for the benefit of accurate high-resolution prediction. In particular, long skip connections are built between encoder and decoder, which enable low-level features to precisely locate targets~\cite{ronneberger2015u}. However, when the low-level features are gradually involved, the global context in high-level features may degrade, leading to pixel-level inconsistency in dense prediction~\cite{yu2018learning}. 

Both backbone-based and encoder-decoder methods demonstrate that pyramid features are essential for accurate semantic segmentation. Nevertheless, they still have the following significant limitations: 1) backbone-based methods are not able to explicitly leverage the low-level features and usually rely on a heavy pre-trained backbone; 2) encoder-decoder methods directly combine multi-level features only in a bottom-up way, i.e., from high to low level, within the decoder; 3) the distinction between low-level and high-level features is neglected, consequently preventing sufficient exploitation of multi-level features.
High-level features in deep layers contribute more to the semantic category recognition of anatomical structures, while low-level features in shallow layers bring advantages to precise boundary generation for high-resolution prediction~\cite{Lin2017Feature}. How to effectively leverage the characteristics and correlations of multi-level features in FCN remains an open question.

To address this issue, we propose a Decoupled Pyramid Correlation Network (DPC-Net) to enhance multi-level features via attention mechanism. In essence, we pay attention to decouple the characteristics of high- and low-level features, and we explicitly model the dependencies between multi-level features to exploit their complementary representations with minimal overhead. Specifically, we first design a powerful Pyramid Feature Encoder (PFE) to extract multi-level features from input images. We decouple the characteristics of features concerning semantic dimension (i.e., channel) and spatial dimension (i.e., height, width, depth). On top of that, we present two types of attention modules, Spatial Correlation (SpaCor) and Semantic Correlation (SemCor) modules, to recursively measure the correlation of multi-level features. The former adaptively enhance spatial details in high-level features with the guidance of low-level ones. The latter selectively emphasizes global semantic information in low-level features with the guidance of high-level ones. The proposed method is an effective strategy to exploit multi-level features and can be easily generalized to other medical image segmentation methods in an end-to-end manner.

To the best of our knowledge, we are the first to explore pyramid correlation via attention mechanism for liver tumor segmentation. In summary, our significant contributions can be summarized as follows.
\begin{itemize}
	\item We formulate the correlation modeling of pyramid features and decouple the correlation in consideration of their characteristics in both spatial and semantic aspects. It effectively strengthens and incorporates pyramid features by leveraging the characteristics of adjacent level features as guidance. 
	\item We design SpaCor and SemCor modules based on the correlation modeling to refine multi-level features recursively in both top-down and bottom-up ways. The proposed attention modules are lightweight, trainable, and can be easily integrated with other multi-level frameworks in an end-to-end manner.
    \item We propose a novel DPC-Net that incorporates PFE, SpaCor, and SemCor for effective and robust liver tumor segmentation. Comprehensive experimental results on the MICCAI 2017 Liver Tumor Segmentation (LiTS) Challenge demonstrate promising performance of the proposed method.
\end{itemize}

A preliminary version of this work has demonstrated the effectiveness of DPC-Net~\cite{Zhang2020DARN}. This paper formulates the pyramid correlation, analyzes underlying design principles for more effective liver tumor segmentation, and adds more experiments to demonstrate its robustness. 

The remainder of this paper is organized as follows. Section~\ref{sec:relatedwork} reviews the related work. Section~\ref{sec:method} presents the details of the proposed DPC-Net. Section~\ref{sec:experiments} elaborates on the experimental results of the proposed method for the application of liver tumor segmentation. Section~\ref{sec:discussion} presents the discussion and Section~\ref{sec:conclusion} concludes the study.

\section{Related Work}
\label{sec:relatedwork}
In the following sections, we review recent advances in semantic segmentation and attention mechanism, and their applications in medical image segmentation.
\subsection{Semantic Segmentation}
As discussed above, FCN for semantic segmentation can be broadly classified into two categories: backbone-based and encoder-decoder methods. Backbone-based methods are built upon a deep backbone network with or without simple up-sampling operations~\cite{Chen2016DeepLab}. 
For example, both Deeplab~\cite{Chen2016DeepLab} and PSPNet~\cite{Zhao2017Pyramid} employ a pre-trained ResNet~\cite{he2016deep} or Xception~\cite{chollet2017xception} as a backbone for feature extraction and a spatial pooling module~\cite{Kaiming2014Spatial} or pyramid pooling~\cite{Zhao2017Pyramid} is proposed to probe multi-level contextual information. Although atrous convolution expands the field of view without down-sampling, it leads to a loss of local details and causes grid issue~\cite{wang2018understanding}. Meanwhile, the low-level features in shallow layers of the backbone are neglected, which restricts the feature reuse~\cite{huang2017densely} and hinders the generation of sharp predictions~\cite{lin2017refinenet}. 

Encoder-decoder methods apply an encoder and a decoder to establish symmetric architecture, aiming for predicting accurate high-resolution segmentation. For example, U-Net~\cite{ronneberger2015u} builds symmetric contracting and expanding paths for microscopy image segmentation. In particular, U-Net enables long skip connections between encoder and decoder to involve low-level features in the precise localization of targets. Refinenet~\cite{lin2017refinenet}, Tiramisu~\cite{jegou2017the}, and GCN~\cite{peng2017large} elaborate on sophisticated decoders that utilize low-level information to help high-level features restore sharp and dense prediction. Nevertheless, the semantic information in high-level features fades as more and more low-level features are connected, resulting in pixel-level inconsistency~\cite{yu2018learning}.

CT images have three dimensions to present the anatomical structures' spatial relation. Existing $2$D methods are not able to fully leverage the 3D information and thus get limited performance. A bunch of variants of FCN that can manage 3D information is proposed~\cite{cicek20163d, milletari2016v, tan2021automatic}. For example, nnU-Net~\cite{isensee2020nnu} elaborates on an adaptive pipeline to automatically design pre-processing, architecture, and post-processing according to the size, resolution, and modality of medical images. SequentialSegNet~\cite{zhang2018sequentialsegnet} combines sequential model with convolution network to exploit both inter- and intra-slice information for multi-organ segmentation. AH-Net~\cite{liu20183d} introduces anisotropic convolutional blocks into the $3$D network for learning representations from anisotropic voxels. H-DenseUNet~\cite{li2018h} incorporates $2$D with $3$D DenseUNet to probe both intra-slice and inter-slice spatial information for better liver tumor segmentation and adopts dense connections to reinforce the information flow. LW-HCN~\cite{zhang2019light} exploits both depth-wise and spatio-temporal separate factorization to reduce trainable parameters, improving the segmentation performance. Though designed exquisitely, most of them rely on a heavy pre-trained backbone, which is complicated and inconvenient in practice. Moreover, they are still facing the low efficiency and high computation burden of $3$D convolution~\cite{li2018h}. Overall, the issues of both backbone-based and encoder-decoder methods remain.

\subsection{Attention Mechanism}
The attention mechanism can capture global and local dependencies, suppress irrelevant noise, and has been widely applied in practice~\cite{vaswani2017attention, jetley2018learn, woo2018cbam}. For instance, Chen et al.~\cite{chen2016attention} propose to weigh the multi-scale features at each pixel location softly. Li et al.~\cite{li2018pyramid} combine attention mechanisms and spatial pyramid to extract particular dense features for semantic segmentation. Fu et al.~\cite{fu2019dual} adaptively integrate local features with their global dependencies for scene parsing. 

Meanwhile, attention modules have been increasingly applied in the field of medical image segmentation. Gu et al.~\cite{gu2020ca} propose a comprehensive attention FCN that incorporates three kinds of attention mechanisms to increase the accuracy and explainability of segmentation. Despite achieving exceptional results, the model draws on 2D convolution and is limited to 3D medical image segmentation. Wang et al.~\cite{wang2020non} design a global aggregation block that leverages self-attention to embed global dependency. Roy et al.~\cite{Roy2018Concurrent} introduce concurrent spatial- and channel-wise squeeze and excitation blocks. Cheng et al.~\cite{cheng2020Fully} employ non-local operations between encoder and decoder. Nevertheless, non-local operation brings a heavy computation burden and forces input size to be restricted due to excessive GPU memory consumption, which is especially critical for 3D segmentation. Besides, these methods only capture the dependencies within the features, while little attention is paid to cross-level correlation. Schlemper et al.~\cite{schlemper2019attention} integrate attention gates in U-Net that learn to suppress irrelevant regions for pancreas segmentation. Wang et al.~\cite{wang2019deep} propose a 3D attention-guided network that harnesses spatial contexts across deep and shallow layers, carrying out precise prostate segmentation from transrectal ultrasound, especially for ambiguous boundaries in the images. Liu et al.~\cite{liu2021spatial} present a SSF-Net that extracts side-outputs at each convolutional block and make full use of them by feature fusion blocks for liver and tumor segmentation. However, they only focus on spatial attention, while the semantic information in the channel dimension of features is neglected. Unlike the previous arts, we adopt attention mechanisms to exploit the correlation among pyramid features in consideration of semantic and spatial dimensions for effective liver tumor segmentation.

\section{MATERIALS AND METHODS}
\label{sec:method}
\begin{figure}[!tp]
\centering
\includegraphics[width=0.9\linewidth]{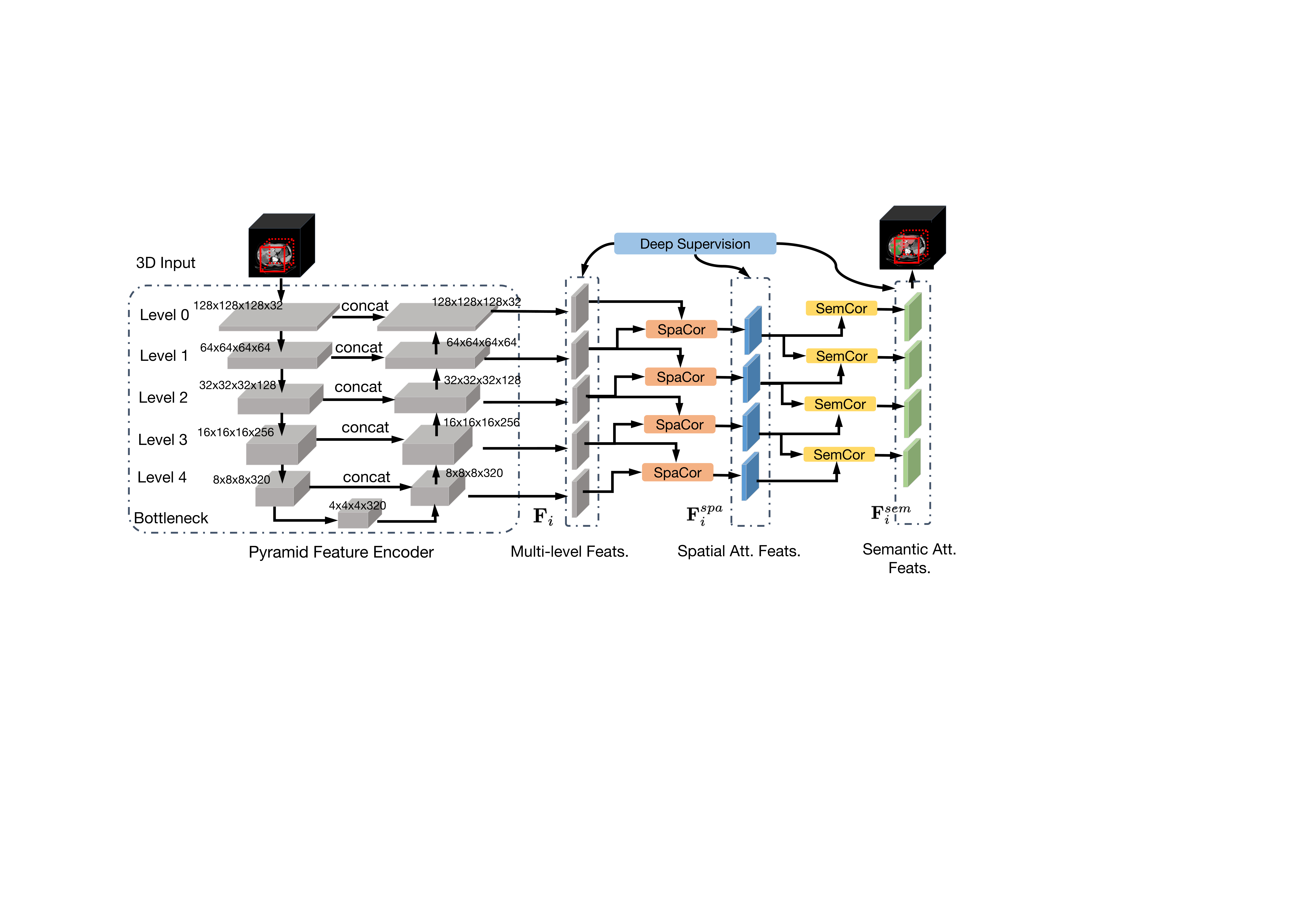}	
\caption{The schematic illustration of the proposed Decoupled Pyramid Correlation Network (DPC-Net), composed of a Pyramid Feature Encoder (PFE), a Spatial Correlation (SpaCor) module, and a Semantic Correlation (SemCor) module. First, PFE extracts multi-level features maps from a 3D patch of the input CT volume (red cube). Then, SpaCor and SemCor modules further refine the multi-level feature maps concerning semantic and spatial dimensions, respectively. The final loss function is a weighted combination of the losses from multi-level, spatial attention, and semantic attention features.
}
\label{fig:overview}
\end{figure}

In this section, we first introduce the dataset and preprocessing procedure and then formulate the modeling of pyramid correlation by attention mechanism. Moreover, we elaborate on the design of DPC-Net consisting of the PFE, the SpaCor module, and the SemCor module. The overview of DPC-Net is illustrated in Figure~\ref{fig:overview}. Also, we explain how these modules specifically handle the aggregation of pyramid features.

\subsection{Dataset and Preprocessing}
The LiTS dataset contains $131$ and $70$ contrast-enhanced $3$D abdominal CT images for training and testing, respectively. The dataset was acquired by different scanners and protocols from six different clinical sites, with a largely varying in-plane resolution from $0.55$ mm to $1.0$ mm and slice spacing from $0.45$ mm to $6.0$ mm. Segmentations of both liver and tumor are included.

\begin{figure}[!tp]
\centering
\includegraphics[width=0.6\linewidth]{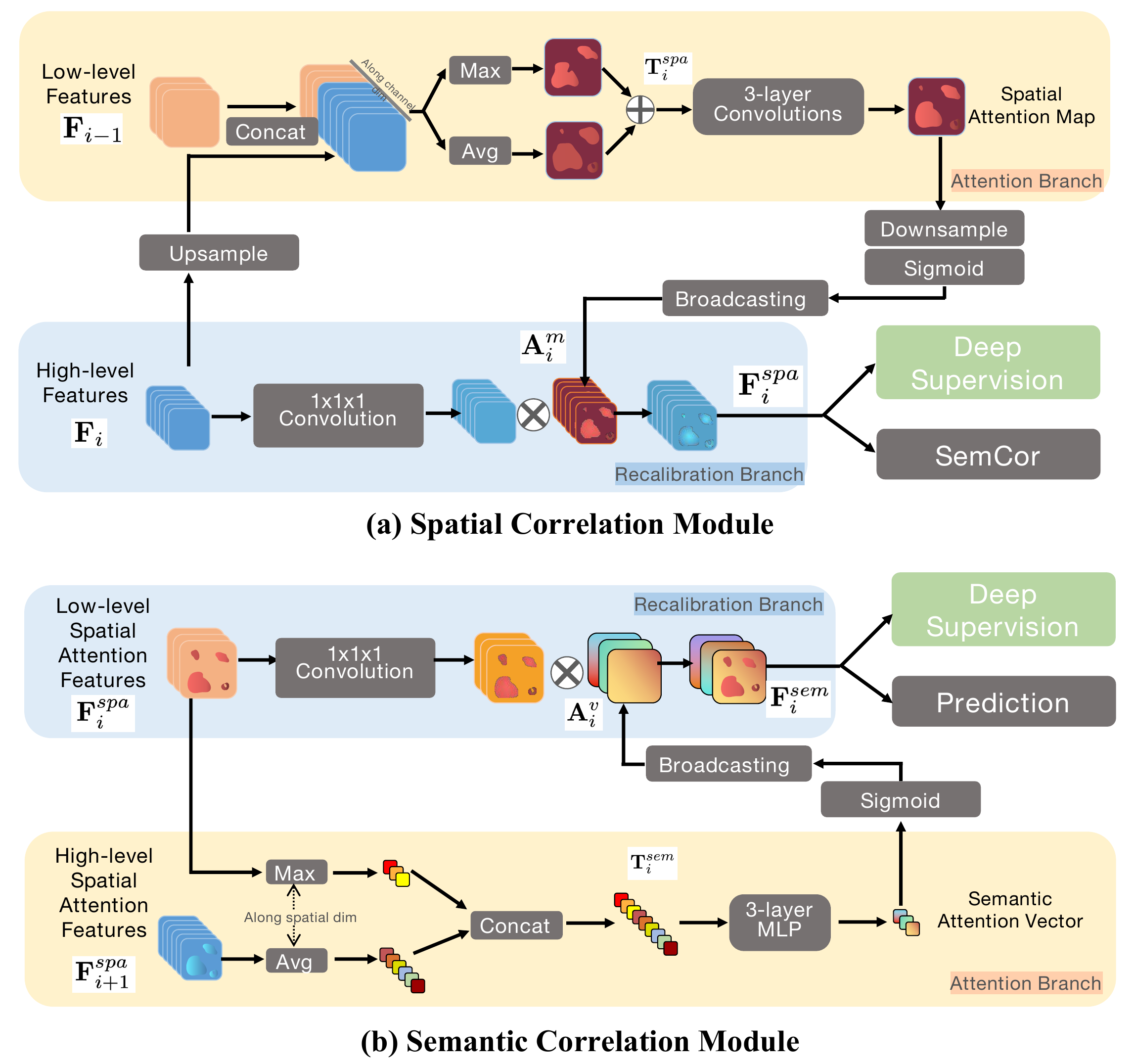}
\caption{Spatial Correlation (SpaCor) and Semantic Correlation (SemCor) Modules. SpaCor refines $\mathbf{F}_{i}$ to output spatial attention features $\mathbf{F}^{spa}_{i}$ with an \emph{attention map} $\mathbf{A}^m_{i}$ derived from low-level features $\mathbf{F}_{i-1}$ while SemCor refines low-level $\mathbf{F}_i$ with an \emph{attention vector} $\mathbf{A}^v_i$ derived from high-level spatial $\mathbf{F}^{spa}_{i+1}$, to generate semantic attention features $\mathbf{F}^{sem}_i$.
}
\label{fig:modules}
\end{figure}

The Hounsfield Unit (HU) value of raw CT volumes is clipped into $[0.5\%, 99.5\%]$ of the initial HU value and normalized with the $Z$-score normalization. The anisotropic voxel spacing is normalized to $(2.47, 1.90, 1.90)$. Considering the limitation of GPU memory, we crop the CT volume into patches with a size of $128 \times 128 \times 128$ pixels. During training, $2/3$ of patches are from random locations within a CT volume, while $1/3$ of patches are guaranteed to contain one of the foreground classes that are present in the CT volume. During inference, the CT images are predicted with a sliding window approach, in which the window size equals the patch size used during training. Adjacent predictions overlap by half of the size of a patch. We apply the online data augmentation during training, including elastic deformations, random scaling, random rotation, and random flipping.

\subsection{Pyramid Correlation Modeling}
\label{sec:overview}
The input image can be embedded as a set of multi-level, i.e. pyramid, features $\mathbf{F} = \{\mathbf{F}_i|i \in [0, L-1]\}$ with different resolutions by CNN, where $\mathbf{F}_i \in \mathbb{R}^{C_i \times D_i \times H_i \times W_i}$ is the features with corresponding scales from the $i$th level and $N$ is the total number of levels. A generic correlation between $\mathbf{F}_i$ and $\mathbf{F}$ can be defined as
\begin{equation}
	\begin{aligned}
	\mathbf{F}^{att}_i=\frac{1}{\mathcal{C}(f(\mathbf{F}_i, \mathbf{F}))} f\left(\mathbf{F}_i, \mathbf{F}\right) g\left(\mathbf{F}_i\right),
	\end{aligned}
	\label{static}
\end{equation}
where $\mathbf{F}^{att}_i$ is the attention feature at the $i$th level, presenting the most informative feature from this level; $f$ is a pair-wise function that represents the relationship of $\mathbf{F}_i$ and $\mathbf{F}$, $g$ is a unary function that generates a mapping of $\mathbf{F}_i$, and $\mathcal{C}$ is a normalization factor. 
Although the pyramid features set encodes both the low-level spatial details and high-level semantic information of the targets, it also inevitably incorporates noise from the shallow layers and loses certain subtle parts of the targets due to the coarse features from deep layers. The low-level features have high resolutions and detailed information, while the high-level features have low resolutions but with richer semantic information.

To fully exploit the characteristics and correlation among pyramid features, we first decouple the characteristics of multi-level features considering spatial and semantic dimensions. It is based on the hypothesis that cross-channel and spatial correlations within features are sufficiently decoupled~\cite{chollet2017xception} and prefer a decoupled processing. Then we build the correlation between adjacent levels, which is formulated as 
\begin{equation}
	\begin{aligned}
	\mathbf{F}^{spa}_i=\frac{1}{\mathcal{C}(f_{spa}\left(\mathbf{F}_i, \mathbf{F}_{i-1}\right))} f_{spa}\left(\mathbf{F}_i, \mathbf{F}_{i-1}\right) g_{spa}\left(\mathbf{F}_i\right),
	\end{aligned}
	\label{SpaCor}
\end{equation}
\begin{equation}
	\begin{aligned}
	\mathbf{F}^{sem}_i=\frac{1}{\mathcal{C}(f_{sem}\left(\mathbf{F}_i, \mathbf{F}_{i+1}\right))} f_{sem}\left(\mathbf{F}_i, \mathbf{F}_{i+1}\right) g_{sem}\left(\mathbf{F}_i\right),
	\end{aligned}
	\label{SemCor}
\end{equation}
where similarly $f_{spa}$ and $f_{sem}$ measure the correlations between adjacent levels with respect to spatial and semantic dimensions respectively, and $g_{spa}$ and $g_{sem}$ are mappings in spatial and semantic dimensions respectively. We wrap the pyramid correlation in Eq.(\ref{SpaCor}) and Eq.(\ref{SemCor}) to a SpaCor and a SemCor modules. They recursively generate the corresponding spatial attention features $\mathbf{F}^{spa} = \{\mathbf{F}_i^{spa}|i \in [1, L)\}$ and semantic attention features $\mathbf{F}^{sem} = \{\mathbf{F}_i^{sem}|i \in [0, L-1)\}$ from the $i$th layer. The SpaCor strengthens high-level features in spatial dimensions (i.e., depth, height, and width dimensions) while the SemCor enhances low-level features in semantic dimension (i.e., channel dimension).

As shown in Figure~\ref{fig:overview}, both SpaCor and SemCor are trainable modules integrated into multi-level architectures and thus can be jointly optimized in an end-to-end manner. Moreover, the extracted semantic and spatial attention features are further enhanced by deep supervision~\cite{dou20173d} on different levels. By capturing the correlations from each layer with the proposed modules, our network learns to select more discriminative features from the image for accurate and robust 3D medical image segmentation. In the following sections, we will elaborate on the detailed design of PFE, SpaCor, and SemCor.

\subsection{Pyramid Feature Encoder}
\label{sec:pfe}
The proposed PFE is a variant of U-Net, and we extend it to embed the raw CT volume as pyramid features. Concretely, we construct a pair of hierarchical contracting and expanding paths and a bottleneck layer, as shown in Figure~\ref{fig:overview}. Every level consists of two cascaded convolution blocks, and each block is composed of a convolution, a normalization, and an activation. A down-sampling operation is added after each layer of the contracting path, and an up-sampling operation is added before each layer of expanding path. In contrast to U-Net~\cite{ronneberger2015u}, the Batch Normalization~\cite{ioffe2015batch} operations after the convolution layers are replaced with Instance Normalization~\cite{ulyanov2016instance} operations, due to the small batch size for training. The normalized features are then activated by a leaky rectified linear unit (LeakyReLU)~\cite{maas2013rectifier} to prevent the neurons from diminishing during the training process. Moreover, the pooling layers are replaced with strided convolution layers. Consequently, the down-sampling operations are learnable. Likewise, the up-sampling operations are transposed convolutions. Afterward, the pyramid features from each layer of the decoder are obtained by using the deep supervision mechanism~\cite{lee2015deeply} that imposes the supervision signals to multiple layers. The deep supervision mechanism reinforces the propagation of gradients flows within the 3D network and, hence, helps learn more representative features~\cite{dou20173d}. It is worthy to note that most FCN can be extended as a pyramid feature encoder mentioned above with minor revision and integrated with the proposed attention modules.

\subsection{Spatial Correlation Module}
\label{sec:SpaCor}

Although the feature maps at deep layers can capture the highly semantic information to indicate the targets' location, they may lose the fine details of the target boundaries. To this end, we design the SpaCor module to adaptively emphasize substantial spatial clues with the guidance of relative low-level features in a top-bottom manner (i.e., from low- to high-level features).

As shown in Figure~\ref{fig:modules} (top), the SpaCor module takes $\mathbf{F}_i$ and $\mathbf{F}_{i-1}$ as inputs and outputs $\mathbf{F}_i^{spa}$. Before that, $\mathbf{F}_i$ is up-sampled to match the resolution of $\mathbf{F}_{i-1}$ and down-sampled to its original size after going through SpaCor. It consists of an attention branch and a recalibration branch. 

Specifically, the attention branch regresses an \emph{attention map} $\mathbf{A}_{i}^m \in \mathbb{R}^{D_i \times H_i \times W_i}$ for each level, which indicates the spatial importance of $\mathbf{F}_{i-1}$ for each $\mathbf{F}_{i}$. The attention map $\mathbf{A}_{i}^m$ is given by
\begin{equation}
	\begin{aligned}
	\mathbf{A}_{i}^m = f_{spa}([\mathbf{F}_{i}, \mathbf{F}_{i-1}]; \theta^{spa}_i),
	\end{aligned}
\end{equation}
where $[\cdot, \cdot]$ is concatenation, $f_{spa}$ measures the spatial correlation of $\mathbf{F}_{i}$ and $\mathbf{F}_{i-1}$, and $\theta^{spa}_i$ represents the trainable parameters of $f_{spa}$ and can be optimized with the network in an end-to-end manner. 
As SpaCor focuses on the spatial correlation, a spatial abstraction $\mathbf{T_i^{spa}}$ is derived from the concatenation of $\mathbf{F}_{i}$ and $\mathbf{F}_{i-1}$ by average and max squeeze on channel dimension. $f_{spa}$ is a $3$-layer convolutional networks due to the nature of convolution for spatial information extraction. 
Formally, $f_{spa}$ is defined as
\begin{equation}
	\begin{aligned}
	f_{spa}(\mathbf{X}) = \mathbf{W}^T_2\phi_{1}(\mathbf{W}^T_1 \mathbf{T}_i^{spa}), \\
	\mathbf{T}_i^{spa} = {Avg}_{spa}(\mathbf{X}) + {Max}_{spa}(\mathbf{X}),
	\end{aligned}
\end{equation}
where $\mathbf{W^T}$ are the weights in the corresponding convolutional layers, and $\phi$ are LeakyReLUs.

Then, in the recalibration branch, we multiply the \emph{attention map} $\mathbf{A}_m$ with $\mathbf{F}_{i}$ in an element-by-element manner to recalibrate the features in $\mathbf{F}_{i}$. Finally, the spatial attention feature maps $\mathbf{F}_{1}^{spa}$ are calculated by
\begin{equation}
	\begin{aligned}
	\mathbf{F}_{i}^{spa} = \sigma(\mathbf{A}_{i}^m) \otimes g_{spa}(\mathbf{F}_{i}),
	\end{aligned}
\end{equation}
where the \emph{attention map} $\mathbf{A}_i^m$ is broadcasted along the channel dimension, $\sigma(x) = \frac{1}{1+e^{x}}$ is a Sigmoid function that normalizes each element of $\mathbf{A}_i^m$ into $(0, 1)$, $g_{spa}$ is a $1\times1\times1$ convolutional layer, and $\otimes$ denotes the element-wise multiplication. 

In general, SpaCor module leverages low-level features as spatial guidance to refine high-level features to involve more discriminative details of targets' boundaries.

\subsection{Semantic Correlation Module}
\label{sec:SemCor}

The features from shallow layers contain the detailed information of targets and features of irrelevant regions. To refine the features from each layer, we present the SemCor module to enhance discriminative semantic response with the guidance of relative high-level features in a bottom-up manner (i.e., from high- to low-level features). It can generate the semantic attention features by utilizing adjacent high-level features with abundant category information to weigh low-level information to select distinct class-specific regions (i.e., liver and tumor).

As illustrated in Figure~\ref{fig:modules} (bottom), the SemCor module takes $\mathbf{F}^{spa}_i$ and $\mathbf{F}^{spa}_{i+1}$ as inputs and outputs $\mathbf{F}_i^{sem}$. Similar to the SpaCor module, the SemCor module is also composed of an attention branch and a recalibration branch. 

Specifically, attention branch takes $\mathbf{F}^{spa}_i$ and $\mathbf{F}^{spa}_{i+1}$ as inputs and then regresses a semantic \emph{attention vector} $\mathbf{A}_i^v \in \mathbb{R}^{C_i}$ as
\begin{equation}
	\begin{aligned}
	\mathbf{A}_i^v = f_{sem}([\mathbf{F}^{spa}_i, \mathbf{F}^{spa}_{i+1}]; \theta^{sem}_i),
	\end{aligned}
\end{equation}
where $\theta^{sem}_i$ denotes the trainable parameters of $f_{sem}$ that measures the semantic correlation of $\mathbf{F}^{spa}_i$ and $\mathbf{F}^{spa}_{i+1}$. The semantic \emph{attention vector} provides global context as a guidance of low-level features to select categorical localization details. 
In contrast to SpaCor, SemCor focuses on the semantic correlation, a semantic abstraction $\mathbf{T}_i^{sem}$ is derived from the concatenation of $\mathbf{F}^{spa}_i$ and $\mathbf{F}^{spa}_{i+1}$ by average and max squeeze on spatial dimension. $f_{sem}$ is a $3$-layer perceptron to model the correlations among semantic channels.
Formally, $f_{sem}$ is defined as
\begin{equation}
	\begin{aligned}
	f_{sem}(\mathbf{X}) = \mathbf{W}^T_2\phi_{1}(\mathbf{W}^T_1 \mathbf{T}_i^{sem}), \\
	\mathbf{T}_i^{sem} = [{Avg}_{sem}(\mathbf{X}), {Max}_{sem}(\mathbf{X})],
	\end{aligned}
\end{equation}
where $\mathbf{W^{T}}$ are the weights in the corresponding fully connected layers, and $\phi$ are LeakyReLUs.

In recalibration branch, we first apply a $1 \times 1 \times 1$ convolution layer to align the channels of $\mathbf{F}^{spa}_i$ with $\mathbf{F}_{i+1}$, and then multiply the \emph{attention vector} $\mathbf{A}_i^v$ with $\mathbf{F}^{spa}_i$ in an element-by-element manner. Finally, in the identity branch, $\mathbf{F}^{spa}_i$ is obtained by
\begin{equation}
	\begin{aligned}
	\mathbf{F}_i^{sem} = \sigma(\mathbf{A}_i^v) \otimes g_{sem}(\mathbf{F}_i),
	\end{aligned}
\end{equation}
where the \emph{attention vector} $\mathbf{A}_i^v$ is broadcasted along the spatial dimension, $\sigma(x)$ is a Sigmoid function, $g_{sem}$ is a fully connected layer, and $\otimes$ denotes the element-wise multiplication. 

This module takes advantage of high-level features to provide guidance information to low-level feature maps in a simple and effective way.

\subsection{Training Strategy}
We employ a hybrid loss that consists of both Cross-Entropy loss and Dice loss. Cross-Entropy loss measures the pixel-wise distribution similarity of ground truth and prediction, while Dice loss measures the overlap and is not sensitive to the size of targets, which alleviates the class imbalance problem~\cite{milletari2016v}. The hybrid loss is formulated as 
\begin{equation}
\begin{aligned}
	\mathcal{L} = \mathcal{L}_{CE} + \mathcal{L}_{\text {Dice }},
\end{aligned}
\end{equation}
\begin{equation}
\begin{aligned}
	\mathcal{L}_{CE} = -\frac{1}{N}\sum_{i=1}^{N}\sum_{c=1}^{C}g_i^c\log{p}_i^c,
\end{aligned}
\end{equation}
\begin{equation}
\begin{aligned}
	\mathcal{L}_{\text {Dice }}=1-\frac{2 \sum_{i=1}^{N} \sum_{c=1}^{C} g_{i}^{c} p_{i}^{c}}{\sum_{i=1}^{N} \sum_{c=1}^{C} g_{i}^{c 2}+\sum_{i=1}^{N} \sum_{c=1}^{C} p_{i}^{c 2}},
\end{aligned}
\end{equation}
where $C$ is the number of classes, and $N$ is the voxel number of one class, $g_i^c$ is a binary indicator if class label $c$ is the correct classification for pixel $i$, and $p_i^c$ is the corresponding predicted probability. 
We employ deep supervision on each level of the PFE, SpaCor, and SemCor. Thus the total loss is defined as the weighted summation of all losses:
\begin{equation}
\begin{aligned}
	\mathcal{L}_{\text {total }} = \sum^{L-1}_{i=0}W_i\mathcal{L}^{PFE}_i + \sum^{L-1}_{i=1}W_i\mathcal{L}^{Spa}_i + \sum^{L-2}_{i=0}W_i\mathcal{L}^{Sem}_i,
\end{aligned}
\end{equation}
where $W_i$ is set as $\frac{1}{2^i*L}$ following nnUNet~\cite{isensee2020nnu}. 


\section{EXPERIMENTS AND RESULTS}
\label{sec:experiments}

\subsection{Evaluation Metrics} 
The metrics employed to evaluate segmentation quantitatively include Dice Similarity Coefficient (DSC), Relative Volume Difference (RVD), Average Symmetric Surface Distance (ASSD) and Hausdorff Distance (HD)\cite{bilic2019the,heimann2009comparison}. 

Let $A$ and $B$ denote the prediction and ground truth. 
DSC and RVD are used to evaluate the area-wise similarity between $A$ and $B$, which are calculated as

\begin{equation}
	\begin{aligned}
	DSC(A,B)=\frac{2|A\cap B|}{|A|+|B|}, RVD(A, B)=\frac{|A|-|B|}{|B|}.
	\end{aligned}
\end{equation}

ASSD measures the average over the shortest distances between the segmented volume and ground truth. 
ASSD is calculated as
\begin{equation}
	\begin{aligned}
	ASSD(A, B)=\frac{1}{|S(A)|+|S(B)|}\left(\sum_{s_{A} \in S(A)} d\left(s_{A}, S(B)\right)+\sum_{s_{B} \in S(B)} d\left(s_{B}, S(A)\right)\right),
	\end{aligned}
\end{equation}
where $S(\cdot)$ is a set of surface points and $d(v, S(A))=\min _{s_{A} \in S(A)}\left\|v-s_{A}\right\|$.

The HD is the longest distance over the shortest distances between the segmented volume and ground truth, which is defined as

\begin{equation}
	\begin{aligned}
	HD(A, B)=\max \left\{\max _{s_{A} \in S(A)} d\left(s_{A}, S(B)\right), \max _{s_{B} \in S(B)} d\left(s_{B}, S(A)\right)\right\}
	\end{aligned}.
\end{equation}

As HD is sensitive to outliers, we use the 95th percentile of the asymmetric HD (HD95) instead of the maximum. 
All evaluation metrics are calculated in $3$D manner. 
A better segmentation has larger values of DSC and smaller absolute values of RVD, ASSD, and HD.

\subsection{Network Setting and Training}
We follow the design of nnUNet \cite{isensee2020nnu} and apply a $5$-level contracting and expanding paths in PFE for pyramid feature generation. In the first level, the number of channels is set as $32$ and are doubled in the next level. The max number of channels is $320$. In PFE, each convolution has a kernel size of $3\times 3\times 3$ except those for down-sampling and up-sampling. The down-samplings are performed by $2\times 2\times 2$ convolution with a stride of $2$ while the up-samplings are conducted by $2\times 2\times 2$ transposed convolution with the same stride of $2$. In SpaCor, each convolution for \emph{attention map} regression has a kernel size of $3\times 3\times 3$, and the number of channels is kept consistent with that of each level. In SemCor, each layer of the perceptron has the same number of neurons with channels at each level. The network is optimized with the Adam optimizer~\cite{kingma2015adam}. The value of the initial learning rate is $0.0001$. We train the network for $1000$ epochs with a batch size of $2$. 

\subsection{Effectiveness of Decoupled Correlation}
In this sub-section, we conduct comprehensive experiments to decompose our DPC-Net and analyze each component's effectiveness. The following experiments are conducted on LiTS training set with 4-fold cross validation. 

\subsubsection{Ablation analysis for Semantic and Spatial Correlation}
U-Net has been widely proved powerful in medical image segmentation. In this experiment, we use the $3$D U-Net~\cite{cicek20163d} as the baseline. For a fair comparison, we extend the $3$D U-Net to generate raw pyramid features in the way described in Section~\ref{sec:pfe}, and then the proposed attention modules are all built on it.

\begin{table}[!tp]
\centering
\caption{Ablation analysis for SemCor and SpaCor on LiTS validation dataset.}
\label{tab:cor}
\resizebox{\textwidth}{!}
{
\begin{tabular}{l|cc|c|c|c|c}
\toprule
\multirow{2}{*}{Method} & \multicolumn{2}{c|}{Pooling}  & \multicolumn{2}{c|}{Liver Tumor} & \multicolumn{2}{c}{Liver} \\ \cline{2-7} 
                        & \multicolumn{1}{c}{Avg} & \multicolumn{1}{c|}{Max}                       & \multicolumn{1}{c|}{DSC [\%]} & \multicolumn{1}{c|}{ASSD [mm]} & \multicolumn{1}{c|}{DSC [\%]} & \multicolumn{1}{c}{ASSD [mm]}  \\ \midrule
Baseline~\cite{cicek20163d} &\texttimes &\texttimes &  51.66 $\pm$ 30.29 & 7.83 $\pm$ 15.89 & 87.85 $\pm$ 8.55 & 5.48 $\pm$ 6.67 \\ \hline
+ SemCor  &\checkmark &\texttimes & 56.19 $\pm$ 29.05 & 7.80 $\pm$ 15.78 & 90.21 $\pm$ 7.48 & 4.80 $\pm$ 5.20 \\ \hline
+ SemCor  &\texttimes &\checkmark & 55.07 $\pm$ 29.06 & 7.77 $\pm$ 15.71 &  90.09 $\pm$ 7.66 & 4.77 $\pm$ 5.31 \\ \hline
+ SemCor  &\checkmark &\checkmark & 57.10 $\pm$ 28.31 & 7.78 $\pm$ 14.63 & 91.04 $\pm$ 6.83 & 3.99 $\pm$ 4.66  \\ \hline
+ SpaCor &\checkmark &\texttimes & 58.02 $\pm$ 28.33 & 7.32 $\pm$ 15.31  & 91.05 $\pm$ 6.60 & 3.89 $\pm$ 4.33 \\ \hline
+ SpaCor &\texttimes &\checkmark & 57.96 $\pm$ 29.01  & 7.27 $\pm$ 14.65 & 91.32 $\pm$ 6.56 & 3.61 $\pm$ 4.45 \\ \hline
+ SpaCor &\checkmark &\checkmark & \textbf{58.76 $\pm$ 28.32} & \textbf{7.26 $\pm$ 15.31} & \textbf{92.29 $\pm$ 6.53} & \textbf{3.54 $\pm$ 4.53} \\ \bottomrule
\end{tabular}
}
\end{table}

\begin{table}[!tp]
\centering
\caption{Ablation analysis for DPC-Net on LiTS validation dataset.
}
\label{tab:dpcnet}
\resizebox{\linewidth}{!}
{
\begin{tabular}{l|ccc|c|c|c|c}
\toprule
\multirow{2}{*}{Method} & \multicolumn{1}{c}{\multirow{2}{*}{SpaCor}} & \multicolumn{1}{c}{\multirow{2}{*}{SemCor}} & \multicolumn{1}{c|}{\multirow{2}{*}{PFE}} & \multicolumn{2}{c|}{Liver Tumor} & \multicolumn{2}{c}{Liver} \\ \cline{5-8} 
                        & \multicolumn{1}{c}{} & \multicolumn{1}{c}{} & \multicolumn{1}{c|}{}                      & \multicolumn{1}{c|}{DSC [\%]} & \multicolumn{1}{c|}{ASSD [mm]} & \multicolumn{1}{c|}{DSC [\%]} & \multicolumn{1}{c}{ASSD [mm]}  \\ \midrule
Baseline~\cite{cicek20163d} &\texttimes &\texttimes &\texttimes  	&  51.66 $\pm$ 30.29          & 7.83 $\pm$ 15.89 & 87.85 $\pm$ 8.55  & 5.48 $\pm$ 6.67        \\ \hline
DPC-Net       &\checkmark  &\texttimes &\texttimes    & 55.11 $\pm$ 29.88 & 7.26 $\pm$ 10.28 & 88.67 $\pm$ 8.26  & 5.05 $\pm$ 4.46  \\ \hline
DPC-Net       &\texttimes  &\checkmark &\texttimes   & 57.10 $\pm$ 28.31 & 7.78 $\pm$ 14.63 & 91.04 $\pm$ 6.83  & 3.99 $\pm$ 4.66  \\ \hline
DPC-Net       &\checkmark  &\checkmark &\texttimes    & 58.76 $\pm$ 28.32 & 7.26 $\pm$ 15.31        & 92.29 $\pm$ 6.53  & 3.54 $\pm$ 4.53  \\ \hline
DPC-Net       &\texttimes  &\texttimes &\checkmark    &  59.89 $\pm$ 28.12 & 6.23 $\pm$ 12.32   & 92.31 $\pm$ 5.98  & 3.31 $\pm$ 4.02  \\ \hline
DPC-Net       &\checkmark  &\checkmark &\checkmark  & \textbf{65.03 $\pm$ 26.85} & \textbf{5.31 $\pm$ 7.11}        & \textbf{94.91 $\pm$ 4.33}  & \textbf{2.56 $\pm$ 3.53}  \\ \bottomrule

\end{tabular}
}
\end{table}

\begin{figure}[!tp]
\centering
\includegraphics[width=\linewidth]{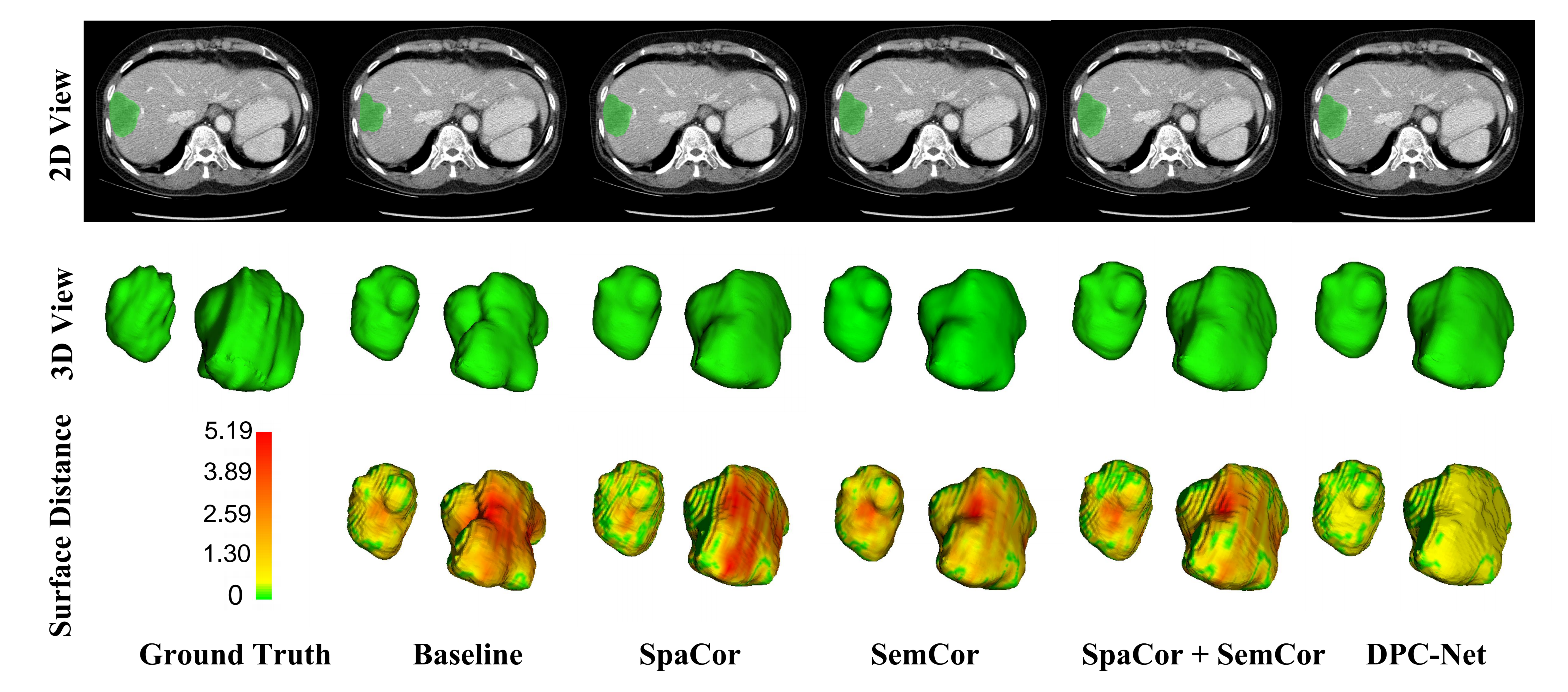}	
\caption{Qualitative results on LiTS dataset. 
The first row shows the $2$D view of the CT images, where tumor segmentations are marked in green. The second row shows the $3$D reconstruction of the segmentations. The third row shows the segmentations' voxel-wise surface distance, where green denotes accurate results while red denotes the other way around. Columns indicate the results derived from ground truth, baseline, SpaCor, SemCor, the combination of SpaCor and SemCor, the proposed DPC-Net, from left to right.
}
\label{fig:lits_ablation}
\end{figure}

To investigate the impact of SemCor, we first evaluate the performance of 3D U-Net as the baseline, and the results are shown in Table.~\ref{tab:cor}. Then we equip the base model with the SemCor. As shown in Table.~\ref{tab:cor}, the SemCor significantly improves the performance of liver tumor segmentation by $5.44\%$ as well as liver segmentation by $3.19\%$ in terms of DSC. Then we compare the impact of using average and max pooling for spatial compression. It reveals that the combination of average and max poolings obtains better results than only each pooling solely.

To investigate the impact of SpaCor, we integrate SpaCor module into the framework. As shown in Table~\ref{tab:cor}, SpaCor further improves the performance of liver tumor segmentation by $1.66\%$ as well as liver segmentation by $1.25\%$. Then, we evaluate two kinds of semantic compression, including only average pooling and the joint of average and max poolings. The results in Table~\ref{tab:cor} reveal that the combination of average and max pooling is more effective.

\subsubsection{Ablation Analysis for DPC-Net} 
To investigate the effectiveness of each component of DPC-Net, we compare $4$ variants of DPC-Net that integrate SemCor, SpaCor, and PFE with the baseline model. Other settings keep consistent in the experiments. 
As shown in Table~\ref{tab:dpcnet}, SpaCor improves the performance of liver tumor segmentation by $3.45\%$ as well as liver segmentation by $0.82\%$ in terms of DSC. Besides, it reduces ASSD of tumor and liver by $0.57$ mm and $0.43$ mm, respectively. SemCor improves the performance of liver tumor segmentation by $5.44\%$ as well as liver segmentation by $3.19\%$ in terms of DSC. Besides, it reduces ASSD of tumor and liver by $0.05$ mm and $1.49$ mm, respectively. The combination of SpaCor and SemCor further improves DSC of liver tumor segmentation by $7.10\%$ as well as liver segmentation by $4.79\%$. Moreover, it reduces the ASSD of tumors and the kidney by $0.57$ mm and $1.86$ mm, respectively. The proposed PFE boost the segmentation performance of tumor and liver by $6.27\%$ and $2.62\%$, respectively. Overall, the proposed DPC-Net obtains the superior improvements by $13.37\%$ and $7.06\%$ in DSC of liver tumor and liver, respectively.
We also analyze the effect of compound loss and the results are shown in Table~\ref{tab:loss}. Dice loss can alleviate the class imbalance problem and achieves better results than Cross Entropy loss, especially for liver tumor segmentation. The combination of Dice and Cross Entropy loss obtains further improvement. SemCor and SpaCor are two independent modules and build the correlation of pyramid features from two different aspects. To explore the effectiveness of the order of SpaCor and SemCor, the results of "SpaCor first" and "SemCor first" on liver tumor segmentation are shown in Table~\ref{tab:attention_tumor}. It is observed that the performance of these two variants are comparable and the difference is not statistically significant with p-value greater than $0.05$.

\begin{table}[!tp]
\centering
\caption{Ablation analysis for loss functions on LiTS validation dataset.
CE and Dice denote Cross Entropy and Dice losses.}
\label{tab:loss}
\begin{tabular}{l|cc|c|c|c|c}
\toprule
\multirow{2}{*}{Method} & \multicolumn{1}{c}{\multirow{2}{*}{CE}} & \multicolumn{1}{c|}{\multirow{2}{*}{Dice}}  & \multicolumn{2}{c|}{Liver Tumor} & \multicolumn{2}{c}{Liver} \\ \cline{4-7} 
                        & \multicolumn{1}{c}{} & \multicolumn{1}{c|}{}                       & \multicolumn{1}{c|}{DSC [\%]} & \multicolumn{1}{c|}{ASSD [mm]} & \multicolumn{1}{c|}{DSC [\%]} & \multicolumn{1}{c}{ASSD [mm]}  \\ \midrule
DPC-Net  &\checkmark &\texttimes & 55.56 $\pm$ 29.75 & 7.81 $\pm$ 15.87 & 91.02 $\pm$ 6.15 & 4.68 $\pm$ 5.15 \\ \hline
DPC-Net  &\texttimes &\checkmark & 57.65 $\pm$ 28.31 & 7.65 $\pm$ 13.33 & 91.65 $\pm$ 6.65 & 3.73 $\pm$ 4.55 \\ \hline
DPC-Net &\checkmark &\checkmark & \textbf{58.76 $\pm$ 28.32} & \textbf{7.26 $\pm$ 15.31} & \textbf{92.29 $\pm$ 6.53} & \textbf{3.54 $\pm$ 4.53} \\ \bottomrule
\end{tabular}
\end{table}

\begin{table}[!tp]
\centering
\caption{Comparison with recent attention-based methods for liver tumor segmentation. $*$ denotes the best results with a p-value less than 0.05.}
\label{tab:attention_tumor}
\resizebox{\linewidth}{!}
{
\begin{tabular}{c|c|c|c|c|c}
\toprule
\multirow{2}{*}{Method} & \multicolumn{1}{c|}{\multirow{2}{*}{Para.}} & \multicolumn{4}{c}{Liver Tumor} \\ \cline{3-6} 
                        & \multicolumn{1}{c|}{} & \multicolumn{1}{c|}{DSC [\%]}  & \multicolumn{1}{c|}{RVD [\%]} & \multicolumn{1}{c|}{ASSD [mm]} & \multicolumn{1}{c}{HD [mm]}  \\ \midrule
Baseline~\cite{cicek20163d} & 22.58M & 51.66 $\pm$ 30.29 & 0.47 $\pm$ 2.19 & 7.83 $\pm$ 15.89 & 37.44 $\pm$ 31.31          \\ \hline
AG~\cite{schlemper2019attention} & +1.02M & 52.41 $\pm$ 31.42 & 0.18 $\pm$ 1.89 & 7.80 $\pm$ 16.01 & 34.42 $\pm$ 28.23  \\ \hline
DAF~\cite{wang2019deep} & +0.68M & 55.81 $\pm$ 29.25 & 0.15 $\pm$ 1.79 & 7.79 $\pm$ 15.21 & 30.17 $\pm$ 30.31  \\ \hline
\makecell[c]{DPC-Net\\(SemCor-first)} & +0.52M & 58.20 $\pm$ 23.58$^*$ & -0.06 $\pm$ 1.05$^*$  & 7.31 $\pm$ 15.33 $^*$& 29.50 $\pm$ 30.21$^*$  \\ \hline
\makecell[c]{DPC-Net\\(SpaCor-first)} & +0.52M & \textbf{58.76 $\pm$ 23.19$^*$} & \textbf{-0.05 $\pm$ 1.09$^*$}& \textbf{7.26 $\pm$ 15.31$^*$} & \textbf{ 28.73 $\pm$ 29.90$^*$}  \\ \bottomrule
\end{tabular}
}
\end{table}

\begin{figure*}[!tp]
\centering
\includegraphics[width=0.9\linewidth]{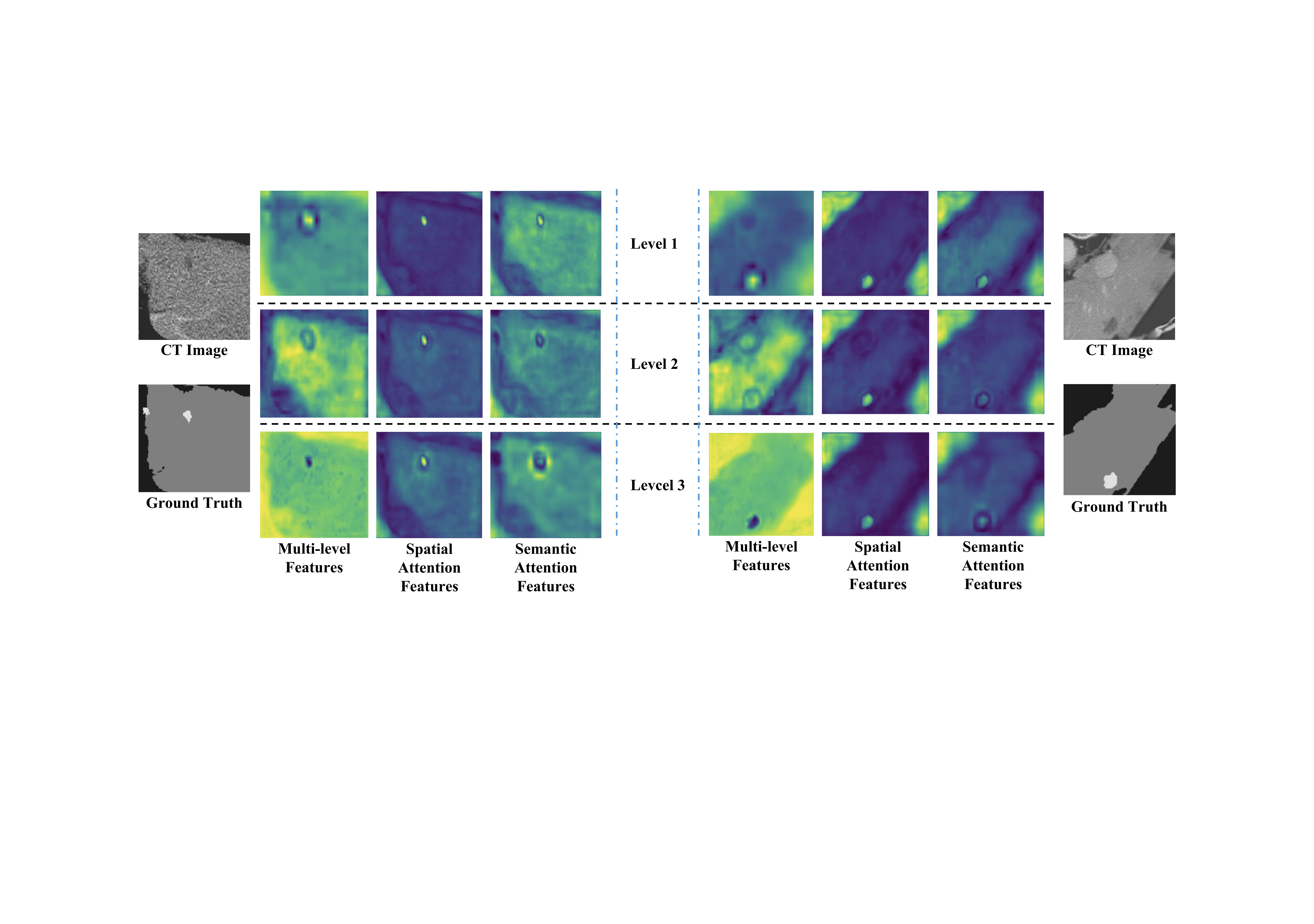}	
\caption{Visualization of two examples to illustrate the effectiveness of the proposed SemCor and SpaCor modules for the feature refinement. The most left and right columns show the input of CT images and their corresponding ground truth. Columns 2 to 4 and 6 to 8 show the corresponding multi-level features, spatial attention features, and semantic attention features from level 1 to 3. It is observed that original multi-level features may suffer from rough localization of liver and tumor and blurred texture of their boundaries. In contrast, semantic and spatial attention features produced by the proposed SpaCor and SemCor are more distinct and discriminative for the better representation of tumors and organs.
}
\label{visualization}
\end{figure*}

\begin{figure}[!htp]
\centering
\includegraphics[width=0.6\linewidth]{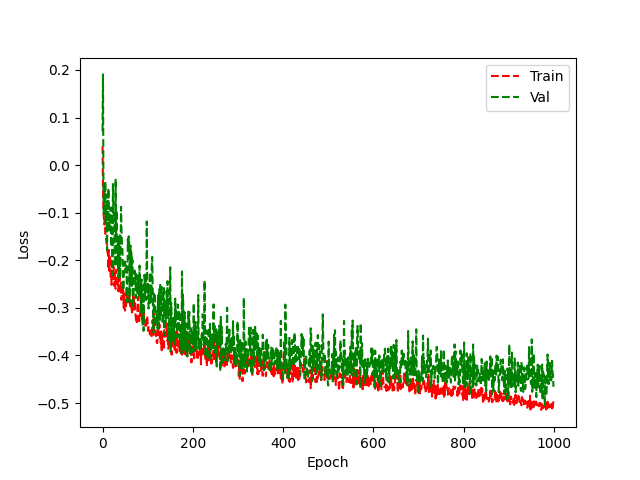}	
\caption{The learning curves of the proposed DCP-Net.}
\label{fig:train_losses}
\end{figure}

The qualitative segmentation results are illustrated in Figure~\ref{fig:lits_ablation}. It is observed that DPC-Net can obtain satisfactory results, especially on the boundaries of tumors, and reduce false positives. The learning curves of training and validation of our method are illustrated in Fig.~\ref{fig:train_losses}. It is observed that the model is converged and not overfitting.
To qualitatively show our attention modules' effects, we visualize the original multi-level and corresponding attention features generated by SpaCor and SemCor, respectively. Figure~\ref{visualization} shows that original multi-level features suffer from the rough localization of liver and tumor and the blurred texture of their boundaries. In contrast, semantic and spatial attention features produced by the proposed SpaCor and SemCor are more distinct and discriminative for robust representation of liver and tumor. Specifically, SpaCor restores the targets' precise location and edge while SemCor suppresses false positives (such as the vessels).

Overall, the proposed SemCor, SpaCor, and PFE obtain consistent performance gain on LiTS datasets, showing the effectiveness and robustness of our method for tumor and organ segmentation.

\subsubsection{Comparison with Attention-based Methods}
\begin{figure}[!tp]
\centering
\includegraphics[width=0.6\linewidth]{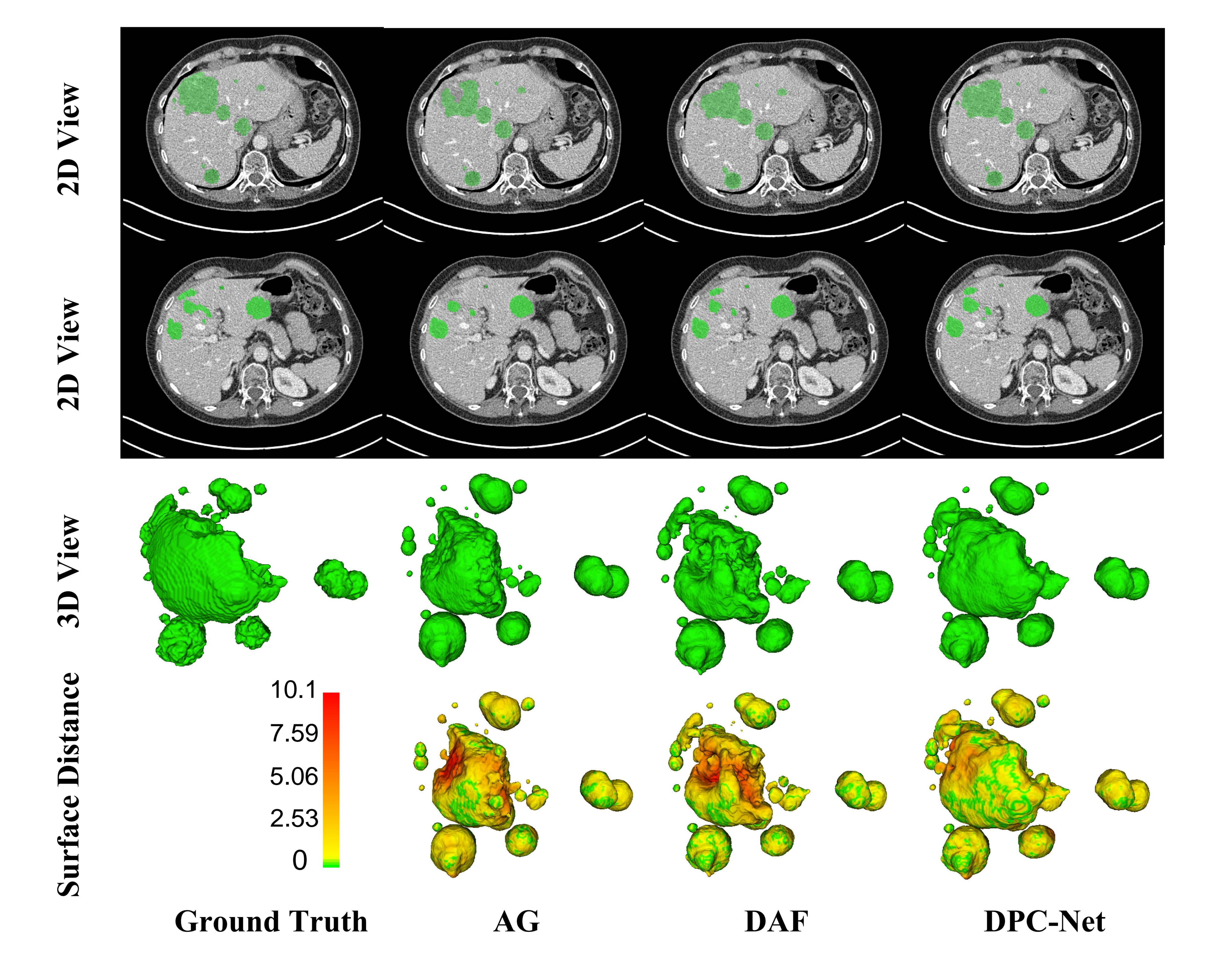}	
\caption{Comparison with attention-based methods on LiTS dataset.
The first two rows show the $2$D view of the same patient's CT images, where tumor segmentations are marked in green. The third row shows the $3$D reconstruction of the segmentations. The fourth row shows the voxel-wise surface distance between the segmentations to ground truth, where green denotes accurate results while red denotes the other way around. Columns indicate the results derived from ground truth, AG, DAF, the proposed DPC-Net, from left to right.
}
\label{fig:lits_attention}
\end{figure}

To the best of our knowledge, there exists no previous art that exploits a pyramid correlation for tumor segmentation. To show our method's effectiveness in modeling multi-level correlation, we compare our method with two recent attention-based methods for medical image segmentation. Schlemper et al.~\cite{schlemper2019attention} design a gated U-Net for pancreas segmentation from CT volumes, while Wang et al.~\cite{wang2019deep} propose a layer-wise attention method for prostate segmentation from a transrectal ultrasound. Since these works are not tailored for tumor segmentation from CT volumes, we do not directly use their tumor segmentation methods. Instead, we adapt their critical ideas to our task with their proposed architecture for a fair comparison. All methods are built upon a $3$D U-Net. We report the performance of our method and these two methods in Table~\ref{tab:attention_tumor}. It is observed that our method outperforms these attention-based methods with a remarkable margin in terms of both DSC and ASSD for both liver and kidney tumor segmentation. Moreover, from the model parameters listed in Table.~\ref{tab:attention_tumor}, we can observe that our method achieves superior results with only additional $0.52$M parameters ($0.43$M in SpaCor and $0.09$M in SemCor) due to the decoupled attention modules. DPC-Net exploits spatial and semantic abstractions, instead of the original feature maps, for attention map estimation to reduce the parameters. Meanwhile, as DPC-Net considers the characteristics of pyramid features from different levels, it still obtains compelling performance. The qualitative results are shown in Figure~\ref{fig:lits_attention}.
We conducted a t-test to to prove the significant improvement of the proposed DPC-Net compared with the other attention-based approaches. In Table~\ref{tab:attention_tumor}, both variants of DPC-Net obtains best results with statistically significant improvement.

\subsection{Comparison with The State-of-the-art Methods}

We also evaluate our method on the testing set of LiTS with ground truths held out by the challenge organizers. Following the evaluation procedures of the LiTS challenge, we evaluate the segmentation performance with DSC and ASSD. We compare DPC-Net with the latest published works of literature that are specially designed for liver tumor segmentation. The nnU-Net~\cite{isensee2020nnu} reports a result that is obtained by cascaded and ensemble strategies. Although these strategies effectively benefit tumor segmentation, they are not practical in clinical practice due to their heavy computation burden. For a fair comparison, we re-implement it with its official code and report a single model's performance on LiTS test set. For other methods, we refer to their results in the corresponding literature and the open leaderboard. It is worthy to note that most methods rely on a pre-trained backbone to obtain promising results. We only trained our model on LiTS training set. It is observed from Table~\ref{tab:lits} that our method equipped with decoupled pyramid attention modules achieves exceeding results on all four metrics for liver tumor segmentation. Besides, DPC-Net also obtains competitive performance for liver segmentation.

\section{Discussion}
\label{sec:discussion}
Recently, deep learning has witnessed an unprecedented improvement in automatic medical image segmentation. Although it achieves considerable improvements on these tasks, accurate segmentation of liver tumor remains very challenging due to the various size, location, and texture of liver tumor, as well as the unclear boundaries. FCN draws success from embedding multi-level features to address these issues. We are motivated to explore pyramid correlations in multi-level features to take advantage of their characteristics fully.

To the best of our knowledge, there is no previous work that explores pyramid correlation via attention mechanism for liver and tumor segmentation. In this work, we investigate the importance of the characteristics and correlation in multi-level features for liver tumor segmentation and propose a DPC-Net to leverage the complementary of pyramid features in both semantic and spatial dimensions. Through the decoupled correlation embedded by SpaCor and SemCor, our method obtains improvement on the segmentation of both liver and tumor, which demonstrates the effectiveness of DPC-Net. Moreover, compared with other advanced attention-based methods~\cite{schlemper2019attention, wang2019deep}, our method is light-weight and probes local and global cross-level correlation effectively. That is crucial in clinical practice when dealing with a considerable amount of medical images. Notably, existing state-of-the-art methods for liver tumor segmentation usually relies on a cascaded strategy and/or pre-training to obtain a promising result~\cite{liu20183d, li2018h, zhang2019light, wang2019volumetric, zabihollahy2021fully}. We show that, equipped with the proposed decoupled correlation, a pure end-to-end 3D network trained from scratch can achieve superior results without complicated processing.

\begin{table}[!tp]
\centering
\caption{Comparison with the state-of-the-art segmentation methods on LiTS testing set.}
\label{tab:lits}
\resizebox{\linewidth}{!}
{
\begin{tabular}{l|c|c|c|c|c|c|c|c|c|c}
\toprule
\multirow{2}{*}{Method}  & \multicolumn{4}{c|}{Liver Tumor} & \multicolumn{4}{c|}{Liver} & \multicolumn{1}{c|}{\multirow{2}{*}{\makecell[c]{Precision at \\ 50\% overlap}}} & \multicolumn{1}{c}{\multirow{2}{*}{\makecell[c]{Recall at \\ 50\% overlap}}}\\ \cline{2-9} 
                        & \multicolumn{1}{c|}{DSC} & \multicolumn{1}{c|}{RVD}  & \multicolumn{1}{c|}{ASSD} & \multicolumn{1}{c|}{HD} & \multicolumn{1}{c|}{DSC} & \multicolumn{1}{c|}{RVD} & \multicolumn{1}{c|}{ASSD} & \multicolumn{1}{c|}{HD} & \multicolumn{1}{c|}{} & \multicolumn{1}{c}{} \\ \midrule
SFF-Net~\cite{liu2021spatial} & 59.2 & - & 1.585 & - & 93.7 & - & 3.678 & - & - & - \\ \hline
AH-Net~\cite{liu20183d} & 63.4 & 0.365 & 1.185 & 6.482 & 96.3 & \textbf{-0.004} & \textbf{1.099} & 2.398 & \textbf{0.468} & 0.301 \\ \hline
H-DenseUNet~\cite{li2018h} & 72.2 & -0.072 & 1.102 & 6.228 & 96.1 & -0.018 & 1.450 & 3.150 & 0.384 & 0.393 \\ \hline
LW-HCN~\cite{zhang2019light} & 73.0 & - & - & - & \textbf{96.5} & - & - & - & - & - \\ \hline
VA-MaskRCNN~\cite{wang2019volumetric} & 74.1 & -0.177 & 1.224 & 6.497 & 96.1 & -0.009 & 1.140 & 2.298 & 0.419 & 0.438 \\ \hline
nnU-Net~\cite{isensee2020nnu} & 74.8 & -0.076 & 1.044 & 6.132 & 96.3 & 0.014 & 1.342 & \textbf{2.134} & 0.437 & \textbf{0.439} \\ \midrule
DPC-Net & \textbf{76.4} & \textbf{-0.063} & \textbf{0.838} & \textbf{5.339} & 96.0 & 0.012 & 1.636 & 4.692 & 0.434 & 0.424 \\ \bottomrule
\end{tabular}
}
\end{table}

In the future, we would like to explore more ways that enable effective pyramid correlation modeling. 
One potential research direction is to use Transformer~\cite{liu2021survey}. 
Furthermore, in clinical practice, multi-phase CT imaging is recommended for better diagnosis of liver tumor~\cite{sun2017automatic}. Another potential research direction is to explore multi-phase tumor segmentation. Most public datasets for liver tumor segmentation only focus on single-phase or modality processing. Multi-phase datasets are highly demanded in this research field.

\section{Conclusion}
\label{sec:conclusion}
This paper proposes a novel DPC-Net that fully exploits multi-level features via an attention mechanism for robust and precise liver tumor segmentation. The proposed DPC-Net employs SemCor and SpaCor modules to refine multi-level features in both top-down and bottom-up ways and incorporate multi-level features by leveraging features from the adjacent level as guidance. Our method achieves promising performance on LiTS dataset, demonstrating the effectiveness for robust and precise liver tumor segmentation. Compared with current attention-based methods, our method is light-weight and effective to model cross-level correlations. Furthermore, the proposed method is generalizable and can be easily integrated with other multi-level frameworks for medical segmentation applications.


\section*{References}
\addcontentsline{toc}{section}{\numberline{}References}
\vspace*{-20mm}





\bibliography{main.bbl}      

\begin{thebibliography}{10}

\bibitem{adelson1984pyramid}
E.~H. Adelson, C.~H. Anderson, J.~R. Bergen, P.~J. Burt, and J.~M. Ogden,
\newblock Pyramid methods in image processing,
\newblock RCA engineer {\bf 29}, 33--41 (1984).

\bibitem{long2015fully}
J.~Long, E.~Shelhamer, and T.~Darrell,
\newblock Fully convolutional networks for semantic segmentation,
\newblock in {\em Proceedings of the IEEE Conference on Computer Vision and
  Pattern Recognition}, pages 3431--3440, 2015.

\bibitem{huang2017densely}
G.~{Huang}, Z.~{Liu}, L.~van~der {Maaten}, and K.~Q. {Weinberger},
\newblock Densely Connected Convolutional Networks,
\newblock in {\em 2017 IEEE Conference on Computer Vision and Pattern
  Recognition (CVPR)}, pages 2261--2269, 2017.

\bibitem{ronneberger2015u}
O.~Ronneberger, P.~Fischer, and T.~Brox,
\newblock U-net: Convolutional networks for biomedical image segmentation,
\newblock in {\em International Conference on Medical Image Computing and
  Computer-Assisted Intervention}, pages 234--241, Springer, 2015.

\bibitem{yu2018learning}
C.~{Yu}, J.~{Wang}, C.~{Peng}, C.~{Gao}, G.~{Yu}, and N.~{Sang},
\newblock Learning a Discriminative Feature Network for Semantic Segmentation,
\newblock in {\em 2018 IEEE/CVF Conference on Computer Vision and Pattern
  Recognition}, pages 1857--1866, 2018.

\bibitem{Lin2017Feature}
T.~Y. Lin, P.~Dollar, R.~Girshick, K.~He, B.~Hariharan, and S.~Belongie,
\newblock Feature Pyramid Networks for Object Detection,
\newblock in {\em 2017 IEEE Conference on Computer Vision and Pattern
  Recognition (CVPR)}, 2017.

\bibitem{Zhang2020DARN}
Y.~{Zhang}, J.~{Tian}, C.~{Zhong}, Y.~{Zhang}, Z.~{Shi}, and Z.~{He},
\newblock DARN: Deep Attentive Refinement Network for Liver Tumor Segmentation
  from 3D CT volume,
\newblock in {\em 2020 IEEE Conference on Pattern Recognition (ICPR)}, 2020.

\bibitem{Chen2016DeepLab}
L.-C. {Chen}, G.~{Papandreou}, I.~{Kokkinos}, K.~{Murphy}, and A.~L. {Yuille},
\newblock DeepLab: Semantic Image Segmentation with Deep Convolutional Nets,
  Atrous Convolution, and Fully Connected CRFs,
\newblock IEEE Transactions on Pattern Analysis and Machine Intelligence {\bf
  40}, 834--848 (2018).

\bibitem{Zhao2017Pyramid}
H.~{Zhao}, J.~{Shi}, X.~{Qi}, X.~{Wang}, and J.~{Jia},
\newblock Pyramid Scene Parsing Network,
\newblock in {\em 2017 IEEE Conference on Computer Vision and Pattern
  Recognition (CVPR)}, pages 6230--6239, 2017.

\bibitem{he2016deep}
K.~{He}, X.~{Zhang}, S.~{Ren}, and J.~{Sun},
\newblock Deep Residual Learning for Image Recognition,
\newblock in {\em 2016 IEEE Conference on Computer Vision and Pattern
  Recognition (CVPR)}, pages 770--778, 2016.

\bibitem{chollet2017xception}
F.~{Chollet},
\newblock Xception: Deep Learning with Depthwise Separable Convolutions,
\newblock in {\em 2017 IEEE Conference on Computer Vision and Pattern
  Recognition (CVPR)}, pages 1800--1807, 2017.

\bibitem{Kaiming2014Spatial}
K.~{He}, X.~{Zhang}, S.~{Ren}, and J.~{Sun},
\newblock Spatial Pyramid Pooling in Deep Convolutional Networks for Visual
  Recognition,
\newblock in {\em European Conference on Computer Vision}, pages 346--361,
  2014.

\bibitem{wang2018understanding}
P.~{Wang}, P.~{Chen}, Y.~{Yuan}, D.~{Liu}, Z.~{Huang}, X.~{Hou}, and
  G.~{Cottrell},
\newblock Understanding Convolution for Semantic Segmentation,
\newblock in {\em 2018 IEEE Winter Conference on Applications of Computer
  Vision (WACV)}, pages 1451--1460, 2018.

\bibitem{lin2017refinenet}
G.~{Lin}, A.~{Milan}, C.~{Shen}, and I.~D. {Reid},
\newblock RefineNet: Multi-path Refinement Networks for High-Resolution
  Semantic Segmentation,
\newblock in {\em 2017 IEEE Conference on Computer Vision and Pattern
  Recognition (CVPR)}, pages 5168--5177, 2017.

\bibitem{jegou2017the}
S.~{Jegou}, M.~{Drozdzal}, D.~{Vazquez}, A.~{Romero}, and Y.~{Bengio},
\newblock The One Hundred Layers Tiramisu: Fully Convolutional DenseNets for
  Semantic Segmentation,
\newblock in {\em 2017 IEEE Conference on Computer Vision and Pattern
  Recognition Workshops (CVPRW)}, pages 1175--1183, 2017.

\bibitem{peng2017large}
C.~{Peng}, X.~{Zhang}, G.~{Yu}, G.~{Luo}, and J.~{Sun},
\newblock Large Kernel Matters - Improve Semantic Segmentation by Global
  Convolutional Network,
\newblock in {\em 2017 IEEE Conference on Computer Vision and Pattern
  Recognition (CVPR)}, pages 1743--1751, 2017.

\bibitem{cicek20163d}
Özgün {Çiçek}, A.~{Abdulkadir}, S.~S. {Lienkamp}, T.~{Brox}, and
  O.~{Ronneberger},
\newblock 3D U-Net: Learning Dense Volumetric Segmentation from Sparse
  Annotation,
\newblock in {\em International Conference on Medical Image Computing and
  Computer-Assisted Intervention}, pages 424--432, 2016.

\bibitem{milletari2016v}
F.~{Milletari}, N.~{Navab}, and S.-A. {Ahmadi},
\newblock V-Net: Fully Convolutional Neural Networks for Volumetric Medical
  Image Segmentation,
\newblock in {\em 2016 Fourth International Conference on 3D Vision (3DV)},
  pages 565--571, 2016.

\bibitem{tan2021automatic}
M.~Tan, F.~Wu, D.~Kong, and X.~Mao,
\newblock Automatic liver segmentation using 3D convolutional neural networks
  with a hybrid loss function,
\newblock Medical Physics {\bf 48}, 1707--1719 (2021).

\bibitem{isensee2020nnu}
F.~Isensee, P.~F. Jaeger, S.~A. Kohl, J.~Petersen, and K.~H. Maier-Hein,
\newblock nnU-Net: a self-configuring method for deep learning-based biomedical
  image segmentation,
\newblock Nature Methods , 1--9 (2020).

\bibitem{zhang2018sequentialsegnet}
Y.~Zhang, X.~Jiang, C.~Zhong, Y.~Zhang, Z.~Shi, Z.~Li, and Z.~He,
\newblock SequentialSegNet: combination with sequential feature for multi-organ
  segmentation,
\newblock in {\em 2018 24th International Conference on Pattern Recognition
  (ICPR)}, pages 3947--3952, IEEE, 2018.

\bibitem{liu20183d}
S.~{Liu}, D.~{Xu}, S.~K. {Zhou}, O.~{Pauly}, S.~{Grbic}, T.~{Mertelmeier},
  J.~{Wicklein}, A.~{Jerebko}, W.~{Cai}, and D.~{Comaniciu},
\newblock 3D Anisotropic Hybrid Network: Transferring Convolutional Features
  from 2D Images to 3D Anisotropic Volumes, 2018.

\bibitem{li2018h}
X.~{Li}, H.~{Chen}, X.~{Qi}, Q.~{Dou}, C.-W. {Fu}, and P.-A. {Heng},
\newblock H-DenseUNet: Hybrid Densely Connected UNet for Liver and Tumor
  Segmentation From CT Volumes,
\newblock IEEE Transactions on Medical Imaging {\bf 37}, 2663--2674 (2018).

\bibitem{zhang2019light}
J.~{Zhang}, Y.~{Xie}, P.~{Zhang}, H.~{Chen}, Y.~{Xia}, and C.~{Shen},
\newblock Light-weight hybrid convolutional network for liver tumor
  segmentation,
\newblock in {\em IJCAI'19 Proceedings of the 28th International Joint
  Conference on Artificial Intelligence}, pages 4271--4277, 2019.

\bibitem{vaswani2017attention}
A.~{Vaswani}, N.~{Shazeer}, N.~{Parmar}, J.~{Uszkoreit}, L.~{Jones}, A.~N.
  {Gomez}, L.~{Kaiser}, and I.~{Polosukhin},
\newblock Attention is All you Need,
\newblock in {\em Proceedings of the 31st International Conference on Neural
  Information Processing Systems}, pages 5998--6008, 2017.

\bibitem{jetley2018learn}
S.~{Jetley}, N.~A. {Lord}, N.~{Lee}, and P.~H.~S. {Torr},
\newblock Learn To Pay Attention,
\newblock in {\em International Conference on Learning Representations}, 2018.

\bibitem{woo2018cbam}
S.~{Woo}, J.~{Park}, J.-Y. {Lee}, and I.~S. {Kweon},
\newblock CBAM: Convolutional Block Attention Module,
\newblock in {\em Proceedings of the European Conference on Computer Vision
  (ECCV)}, pages 3--19, 2018.

\bibitem{chen2016attention}
L.-C. {Chen}, Y.~{Yang}, J.~{Wang}, W.~{Xu}, and A.~L. {Yuille},
\newblock Attention to Scale: Scale-Aware Semantic Image Segmentation,
\newblock in {\em 2016 IEEE Conference on Computer Vision and Pattern
  Recognition (CVPR)}, pages 3640--3649, 2016.

\bibitem{li2018pyramid}
H.~{Li}, P.~{Xiong}, J.~{An}, and L.~{Wang},
\newblock Pyramid Attention Network for Semantic Segmentation.,
\newblock in {\em BMVC}, page 285, 2018.

\bibitem{fu2019dual}
J.~{Fu}, J.~{Liu}, H.~{Tian}, Y.~{Li}, Y.~{Bao}, Z.~{Fang}, and H.~{Lu},
\newblock Dual Attention Network for Scene Segmentation,
\newblock in {\em 2019 IEEE/CVF Conference on Computer Vision and Pattern
  Recognition (CVPR)}, pages 3146--3154, 2019.

\bibitem{gu2020ca}
R.~{Gu}, G.~{Wang}, T.~{Song}, R.~{Huang}, M.~{Aertsen}, J.~{Deprest},
  S.~{Ourselin}, T.~{Vercauteren}, and S.~{Zhang},
\newblock CA-Net: Comprehensive Attention Convolutional Neural Networks for
  Explainable Medical Image Segmentation.,
\newblock IEEE Transactions on Medical Imaging , 1--1 (2020).

\bibitem{wang2020non}
Z.~{Wang}, N.~{Zou}, D.~{Shen}, and S.~{Ji},
\newblock Non-Local U-Nets for Biomedical Image Segmentation.,
\newblock Proceedings of the AAAI Conference on Artificial Intelligence {\bf
  34}, 6315--6322 (2020).

\bibitem{Roy2018Concurrent}
A.~G. Roy, N.~Navab, and C.~Wachinger,
\newblock {\em Concurrent Spatial and Channel Squeeze \& Excitation in Fully
  Convolutional Networks},
\newblock Springer, Cham, 2018.

\bibitem{cheng2020Fully}
J.~Cheng, S.~Tian, L.~Yu, H.~Lu, and X.~Lv,
\newblock Fully Convolutional Attention Network for Biomedical Image
  Segmentation,
\newblock Artificial Intelligence in Medicine {\bf 107}, 101899 (2020).

\bibitem{schlemper2019attention}
J.~{Schlemper}, O.~{Oktay}, M.~{Schaap}, M.~P. {Heinrich}, B.~{Kainz},
  B.~{Glocker}, and D.~{Rueckert},
\newblock Attention gated networks: Learning to leverage salient regions in
  medical images,
\newblock Medical Image Analysis {\bf 53}, 197--207 (2019).

\bibitem{wang2019deep}
Y.~{Wang}, D.~{Ni}, H.~{Dou}, X.~{Hu}, L.~{Zhu}, X.~{Yang}, M.~{Xu}, J.~{Qin},
  P.-A. {Heng}, and T.~{Wang},
\newblock Deep Attentive Features for Prostate Segmentation in 3D Transrectal
  Ultrasound,
\newblock IEEE Transactions on Medical Imaging {\bf 38}, 2768--2778 (2019).

\bibitem{liu2021spatial}
T.~Liu, J.~Liu, Y.~Ma, J.~He, J.~Han, X.~Ding, and C.-T. Chen,
\newblock Spatial feature fusion convolutional network for liver and liver
  tumor segmentation from CT images,
\newblock Medical Physics {\bf 48}, 264--272 (2021).

\bibitem{dou20173d}
Q.~{Dou}, L.~{Yu}, H.~{Chen}, Y.~{Jin}, X.~{Yang}, J.~{Qin}, and P.-A. {Heng},
\newblock 3D deeply supervised network for automated segmentation of volumetric
  medical images,
\newblock Medical Image Analysis {\bf 41}, 40--54 (2017).

\bibitem{ioffe2015batch}
S.~{Ioffe} and C.~{Szegedy},
\newblock Batch Normalization: Accelerating Deep Network Training by Reducing
  Internal Covariate Shift,
\newblock in {\em Proceedings of The 32nd International Conference on Machine
  Learning}, pages 448--456, 2015.

\bibitem{ulyanov2016instance}
D.~{Ulyanov}, A.~{Vedaldi}, and V.~S. {Lempitsky},
\newblock Instance Normalization: The Missing Ingredient for Fast Stylization,
\newblock arXiv preprint arXiv:1607.08022  (2016).

\bibitem{maas2013rectifier}
A.~L. Maas, A.~Y. Hannun, and A.~Y. Ng,
\newblock Rectifier nonlinearities improve neural network acoustic models,
\newblock in {\em Proc. icml}, volume~30, page~3, 2013.

\bibitem{lee2015deeply}
C.-Y. {Lee}, S.~{Xie}, P.~W. {Gallagher}, Z.~{Zhang}, and Z.~{Tu},
\newblock Deeply-Supervised Nets,
\newblock in {\em Proceedings of the Eighteenth International Conference on
  Artificial Intelligence and Statistics}, pages 562--570, 2015.

\bibitem{bilic2019the}
P.~{Bilic} and et~al.,
\newblock The Liver Tumor Segmentation Benchmark (LiTS).,
\newblock arXiv preprint arXiv:1901.04056  (2019).

\bibitem{heimann2009comparison}
T.~{Heimann} and et~al.,
\newblock Comparison and Evaluation of Methods for Liver Segmentation From CT
  Datasets,
\newblock IEEE Transactions on Medical Imaging {\bf 28}, 1251--1265 (2009).

\bibitem{kingma2015adam}
D.~P. {Kingma} and J.~L. {Ba},
\newblock Adam: A Method for Stochastic Optimization,
\newblock in {\em ICLR 2015 : International Conference on Learning
  Representations 2015}, 2015.

\bibitem{wang2019volumetric}
X.~{Wang}, S.~{Han}, Y.~{Chen}, D.~{Gao}, and N.~{Vasconcelos},
\newblock Volumetric Attention for 3D Medical Image Segmentation and Detection,
\newblock in {\em International Conference on Medical Image Computing and
  Computer-Assisted Intervention}, pages 175--184, 2019.

\bibitem{zabihollahy2021fully}
F.~Zabihollahy, A.~N. Viswanathan, E.~J. Schmidt, M.~Morcos, and J.~Lee,
\newblock Fully automated multi-organ segmentation of female pelvic magnetic
  resonance images with coarse-to-fine convolutional neural network,
\newblock Medical physics  (2021).

\bibitem{liu2021survey}
Y.~Liu, Y.~Zhang, Y.~Wang, F.~Hou, J.~Yuan, J.~Tian, Y.~Zhang, Z.~Shi, J.~Fan,
  and Z.~He,
\newblock A Survey of Visual Transformers,
\newblock arXiv preprint arXiv:2111.06091  (2021).

\bibitem{sun2017automatic}
C.~Sun, S.~Guo, H.~Zhang, J.~Li, M.~Chen, S.~Ma, L.~Jin, X.~Liu, X.~Li, and
  X.~Qian,
\newblock Automatic segmentation of liver tumors from multiphase
  contrast-enhanced CT images based on FCNs,
\newblock Artificial intelligence in medicine {\bf 83}, 58--66 (2017).

\end{thebibliography}



\bibliographystyle{medphy.bst}    


\end{document}